\documentclass[final]{cvpr}

\usepackage{times}
\usepackage{epsfig}
\usepackage{graphicx}
\usepackage{amsmath}
\usepackage{amssymb}
\usepackage{comment}

\usepackage[pagebackref=true,breaklinks=true,colorlinks,bookmarks=false]{hyperref}

\def\cvprPaperID{4} 
\def\confYear{CVPR 2021}
\begin{document}

\title{X-MAN: Explaining multiple sources of anomalies in video}

\author{Stanislaw Szymanowicz\\
University of Cambridge\\
{\tt\small sks57@cam.ac.uk}
\and
James Charles\\
University of Cambridge\\
{\tt\small jjc75@cam.ac.uk}
\and
Roberto Cipolla\\
University of Cambridge\\
{\tt\small rc10001@cam.ac.uk}
}

\maketitle

\begin{abstract}
Our objective is to detect anomalies in video while also automatically explaining the reason behind the detector's response. In a practical sense, explainability is crucial for this task as the required response to an anomaly depends on its nature and severity. However, most leading methods (based on deep neural networks) are not interpretable and hide the decision making process in uninterpretable feature representations. In an effort to tackle this problem we make the following contributions: (1) we show how to build interpretable feature representations suitable for detecting anomalies with state of the art performance, (2) we propose an interpretable probabilistic anomaly detector which can describe the reason behind it's response using high level concepts, (3) we are the first to directly consider object interactions for anomaly detection and (4) we propose a new task of explaining anomalies and release a large dataset for evaluating methods on this task.  Our method competes well with the state of the art on public datasets while also providing anomaly explanation based on objects and their interactions.

\end{abstract}


\begin{figure}[h]
\includegraphics[width=\linewidth]{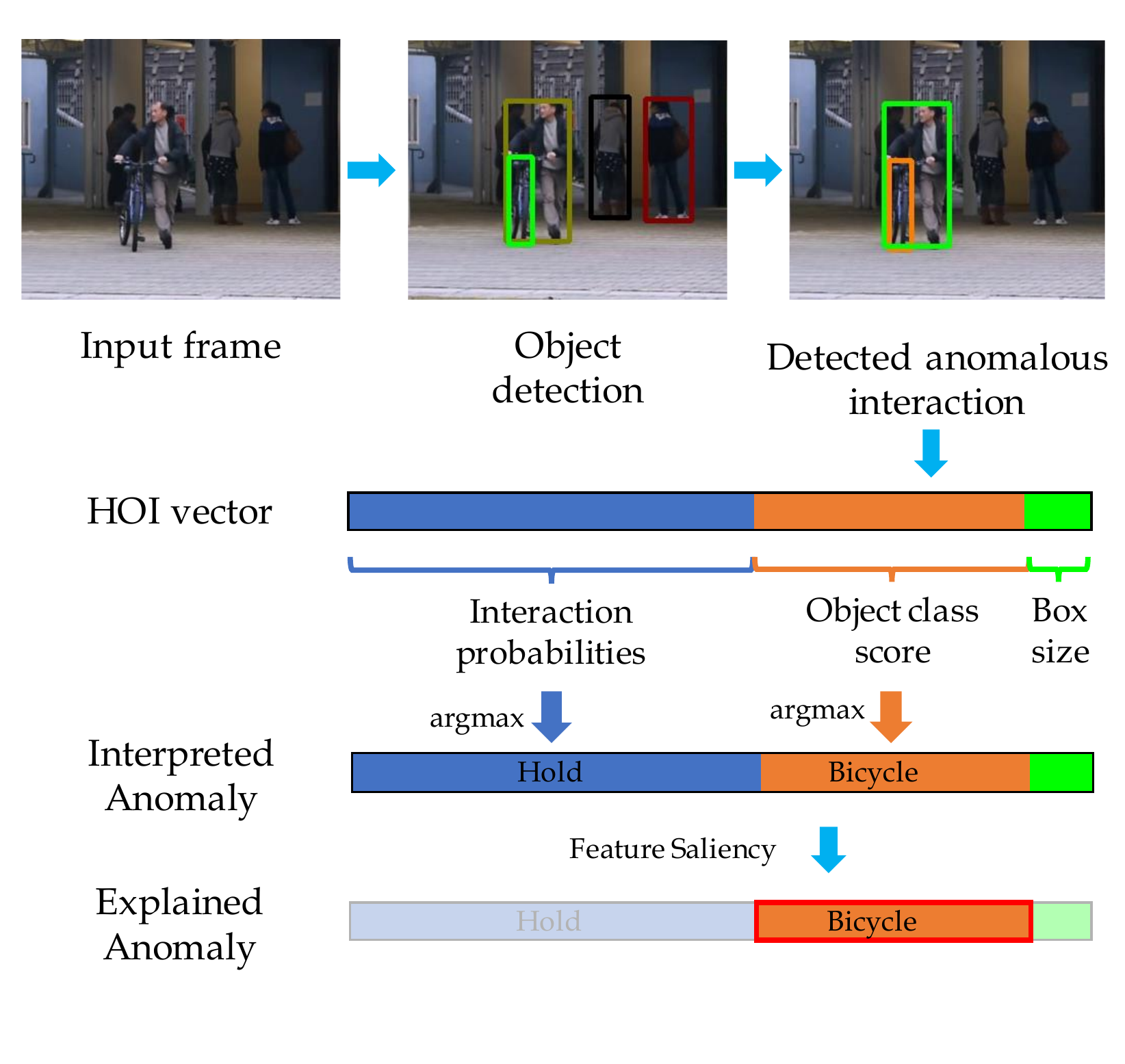}
\caption{\textbf{Overview.} Every human-object pair in a single input frame of a video is encoded into a high-level and interpretable feature vector (HOI vector) based on 
their interactions, object category and person box size. Our method can describe anomalous events in a video by comparing this feature vector to normal ones, highlighting both where in the scene and also where in the feature vector the anomaly occurs. The above example illustrates the detected objects and the detected anomalous human-object interaction: ``holding a bicycle" with the most salient feature for the anomaly being the ``bicycle" object. Note that bicycle holding is anomalous here as previous data suggests it is unusual to observe bicycles from this camera.} \label{paper_summary}
\end{figure}

\section{Introduction}
All over the world there are an ever growing number of surveillance monitoring systems, in streets, in shops and our homes. In the UK, it is estimated there are 500,000 cameras in Greater London alone with an estimate of at least 1 camera per 14 people in the whole of the country~\cite{mccahill2002cctv}. Usually these systems are used for post event analysis or online event detection, in automatic settings object detectors (e.g. people, faces) are typically used. A rather more difficult task is that of detecting anomalies, whereby the scene can be complex with many moving objects and the anomaly itself can only be defined by the environmental context. We also believe anomaly detection should be interpretable, if the model truly understands the scene then it should be able to explain it's response. We address this issue here by considering interpretable object interactions in video, producing an explainable anomaly detection model and building a dataset for evaluating the quality of explanations.

\begin{figure*}[t]
\includegraphics[width=\textwidth]{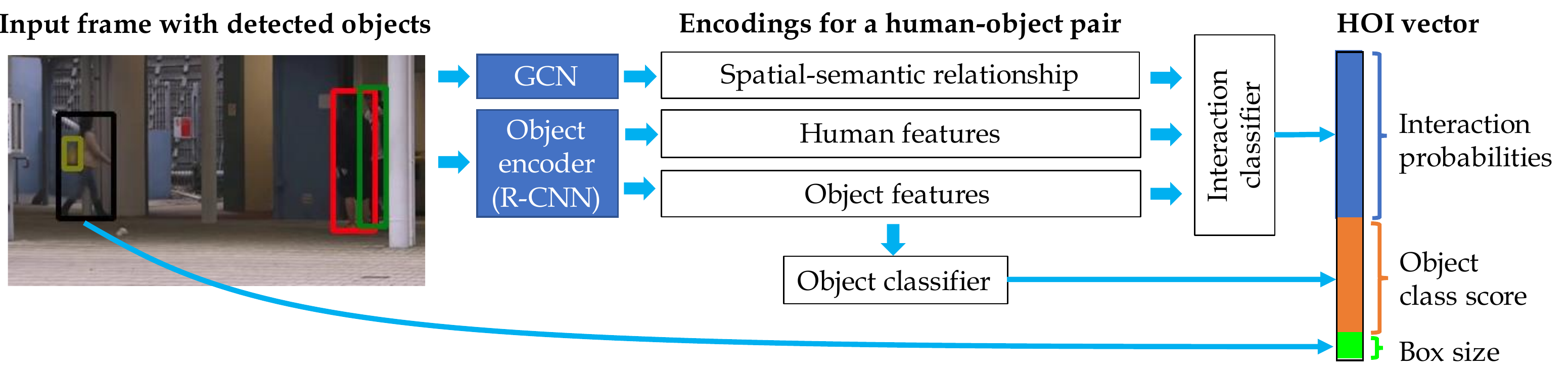}
\caption{\textbf{HOI vector.} All pairs of detected objects (from an R-CNN network) are encoded into (1) a spatial and semantic encoding using a graph convolutional network (GCN), (2) human appearance feature and (3) object appearance feature. Two classifiers ingests these features to produce interaction probabilities and object class scores. These probabilities, together with bounding box size, are concatenated to form the HOI vector. 
} \label{interaction_inference}
\end{figure*}

Being able to both detect \emph{and} explain anomalies has an important application in many types of video monitoring settings. For example, human operators needs to be able to decide on the appropriate response, i.e. simply staying more vigilant if someone is walking in the wrong direction or intervening in person when there is a fight. This is challenging, because (a) the most alarming anomalies involve complex interactions between objects, e.g. people pushing each other next to a road, (b) interpreting the anomaly requires high level understanding of the video context, e.g. is anomalous motion in the scene due to a person running, or a tree blowing in the wind? and (c) the model also has to explain it's response. 

In an effort to address all three of these issues, under the task of explainable anomaly detection, we make the following contributions: (1) we show how to build interpretable feature representations suitable for detecting anomalies with state of the art performance, (2) we propose an interpretable probabilistic anomaly detector which can describe the reason behind it's response using high level concepts, (3) we are the first to directly consider object interactions for anomaly detection and (4) we propose a new task of explaining anomalies and release an accompanying dataset, X-MAN (eXplanations of Multiple sources of ANomalies), the first dataset for evaluating explanations in anomaly detection. Our system (illustrated in Fig.~\ref{paper_summary}) is capable of (i) outputting a high level description of the anomalous event e.g. holding a bicycle, (ii) recovering the most similar normal event e.g. holding a bag and (iii) is also able to identify the most salient features (from a deep neural network) causing the anomaly, e.g. the held bicycle object. 

\section{Related work}

\subsection{Anomaly detection}

A common approach to anomaly detection with deep learning is to predict future `normal' video frames from sparse feature representations~\cite{hasan_2016,luo_2017,wang_2018}, sometimes augmented with memory modules~\cite{park_2020}, and/or optical-flow images~\cite{liu_2018,GANs,Cross_channel_GANs}. Anomalies are detected based on the assumption these models will find it difficult to generate abnormal frames.
An alternative approach is using or learning these feature representations with out of distribution detection algorithms~\cite{hendrycks2018deep,ren2019likelihood,hsu2020generalized}, as can be done for anomaly detection~\cite{mousavi_2015,sabokrou_2015,zhao_2016} or more recently and successfully, applying a one-vs-all cluster classification~\cite{ionescu_2019_cvpr,ionescu_2019_wacv}. However, these methods do not explain the high level content of the scene, the `typical' normal events or explain why an event is abnormal in an interpretable way. Our work is novel in that it uses traditional probabilistic modelling but leverages a deep network's ability to extract high level information about the scene to allow an event to be (1) interpretable and (2) explainable based on a saliency mapping over a learnt feature representation.

\subsection{Anomaly explanation}

To the best of our knowledge~\cite{hinami_2017} is the only work tackling the task of describing the detected 
anomalies. Here, a sparse feature representation of the scene is simultaneously used as input to an anomaly classifier and object, action and attribute detector. The shortcoming of this approach is neglecting the interaction between objects. In many applications the interactions between objects are the crucial source of anomaly, for example it may not be abnormal for a person to jump, but it \emph{is} abnormal for a person to jump over the gates of the underground. Moreover, this approach is still not fully explainable, i.e. input features for detecting anomalies cannot be inspected. Our method uses higher level reasoning than~\cite{hinami_2017} (both high level object categories and human-object interactions) while also justifying the algorithms decision by returning interpretable salient features. Due to the lack of labelled data, there are also difficulties in forming a benchmark on the task of anomaly explanation. To this end, we build such a dataset and show how to make improvements in explaining anomalies by using human-object interactions.

\section{Method}
In Sec.~\ref{HOI_recognition} we outline how we encode the frames of a video into objects and human-object interactions. In Sec.~\ref{anomaly_detection} we detail the  anomaly detection method, and in Sec.~\ref{anomaly_explanation} how we interpret and explain the decisions from the anomaly detector.

\begin{figure*}[t]
\includegraphics[width=\textwidth]{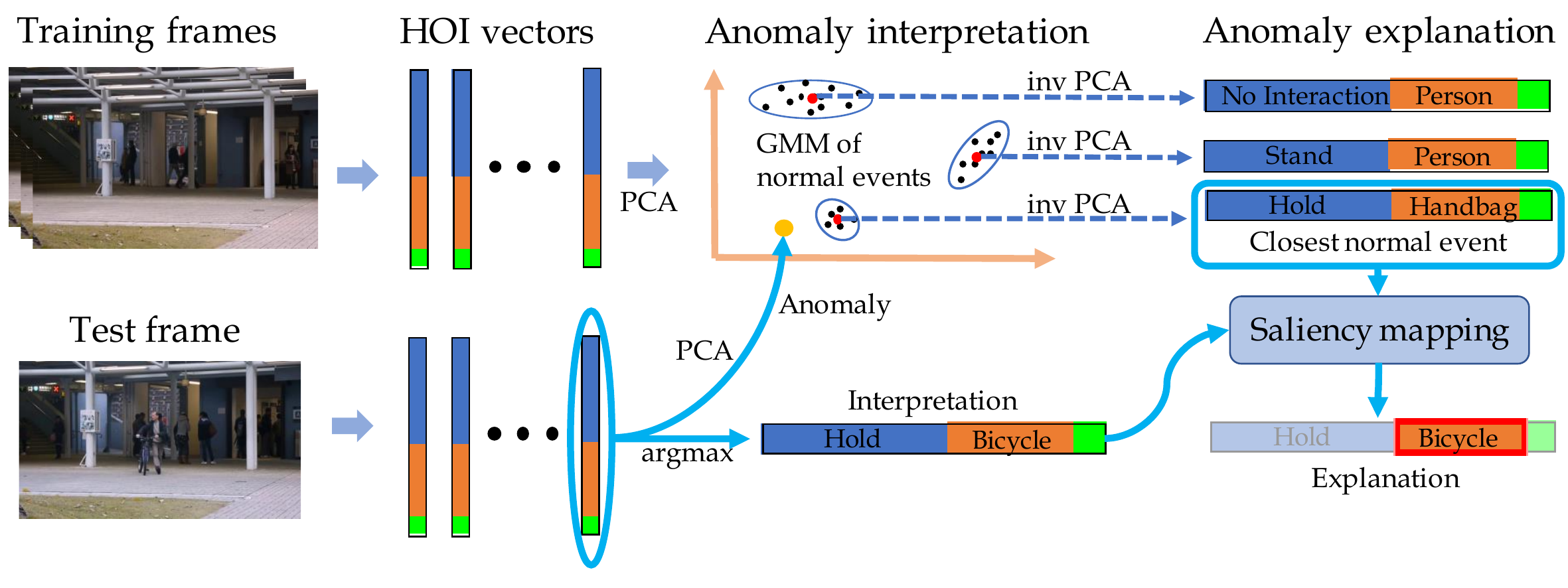}
\caption{\textbf{Method.} Top: training images are pre-processed to obtain  
HOI vectors, which are then used to train a Gaussian Mixture Model.
The estimated means are interpreted as normal events, later used for explanation. Bottom: Test frame detected as anomalous, showing at least one HOI vector has low probability under the GMM.
High-level information in the vector is inspected to interpret the anomaly. Saliency mapping highlights the most salient features by comparison to 
the closest normal HOI vector.
} \label{method}
\end{figure*}

\subsection{Encoding the video frame}\label{HOI_recognition}

Every frame of a video is encoded into a set of human-object interaction vectors (HOI vectors), see Fig.~\ref{interaction_inference}. HOI vectors are interpretable as they are formed from high level interaction and object features. For every human-object pair in the scene a HOI vector is recovered and consists of object class score (one-hot encoding with object class confidence from the object detector in place of the ``1"), the probabilities of each of the interaction classes and person bounding box dimensions. We use Faster R-CNN \cite{ren_frcnn_2015} for object detection, followed by Dual Relation Graph \cite{gao_2020_drg} for Human-Object Interaction recognition. 

\subsection{Anomaly detection}\label{anomaly_detection}

We formalize the anomaly detection as a probabilistic outlier detection problem. The scene is represented by a statistical model of the HOI vectors. An outlier/anomaly is declared when the probability of a HOI vector under this model is below a threshold. 

All HOI vectors from normal (not anomalous) frames are first reduced in dimensionality using PCA. We train a Gaussian Mixture Model (GMM) from these representations of `normal' as in Fig.~\ref{method}(top). This model allows for a probabilistic modelling of the environment. During inference we declare a video frame anomalous if any of it's HOI vectors falls below a threshold probabilty under the GMM Fig.~\ref{method}(bottom). Using our proposed saliency mapping, an interpretable explanation for the model can also be derived, this is detailed further below.

\paragraph{Temporal information} In order to use make use of the temporal information provided by a video we smooth the per-frame anomaly scores with a Gaussian filter.

\subsection{Anomaly explanation}\label{anomaly_explanation}

Our method allows for explanation of abnormalities on three levels: an explanation of the tested event, the closest normal event and the salient features responsible for the deviation. We detail the methods used to explain each of these below:

\paragraph{Tested event.} In our framework, the vector describing an event consists of high-level information: object class, interaction class probabilities and bounding box size. Any event can be explained by a direct inspection of the corresponding HOI vector $\boldsymbol{v}$, as shown in Fig.\ref{method}. PCA is appled to reduce the dimensionality of the HOI vector: $\boldsymbol{x}=W^{T}\boldsymbol{v}$, where $W$ is a weight matrix.

\paragraph{Normal event.} We fit a GMM to the distribution of all PCA reduced HOI vectors from all normal video frames in the training set using $M$ mixture components. The set of means for modes $m \in \{1,\dots,M\}$, denoted $\{\boldsymbol{\mu_m}\}$ represents our set of normal events.

\paragraph{Deviation from normal.} The deviation $\boldsymbol{\Delta}_{m}$ of an anomaly $\boldsymbol{x}$ (PCA-reduced HOI vector) from the closest normal event $\boldsymbol{\mu_m}$ can be measured as a probability of being generated by mixture component $m \in \{1,\dots,M\}$:
\begin{equation}
\begin{split}
p(\boldsymbol{x}|m) & = \frac{1}{2\pi |\Sigma_{m}|}\textrm{exp}\left( -\frac{1}{2}\boldsymbol{\Delta_{m}}^{T} \Sigma_{m}^{-1} \boldsymbol{\Delta_{m}}\right) \\ \boldsymbol{\Delta_{m}} & =\boldsymbol{x}-\boldsymbol{\mu_{m}}
\end{split}
\label{eq:deviation}
\end{equation}
Where $\Sigma_{m}$ is the covariance matrix of $k$ and the quadratic term inside the exponential is a weighted distance of the anomaly from the closest normal event.

\begin{figure*}[t]
\includegraphics[width=\textwidth]{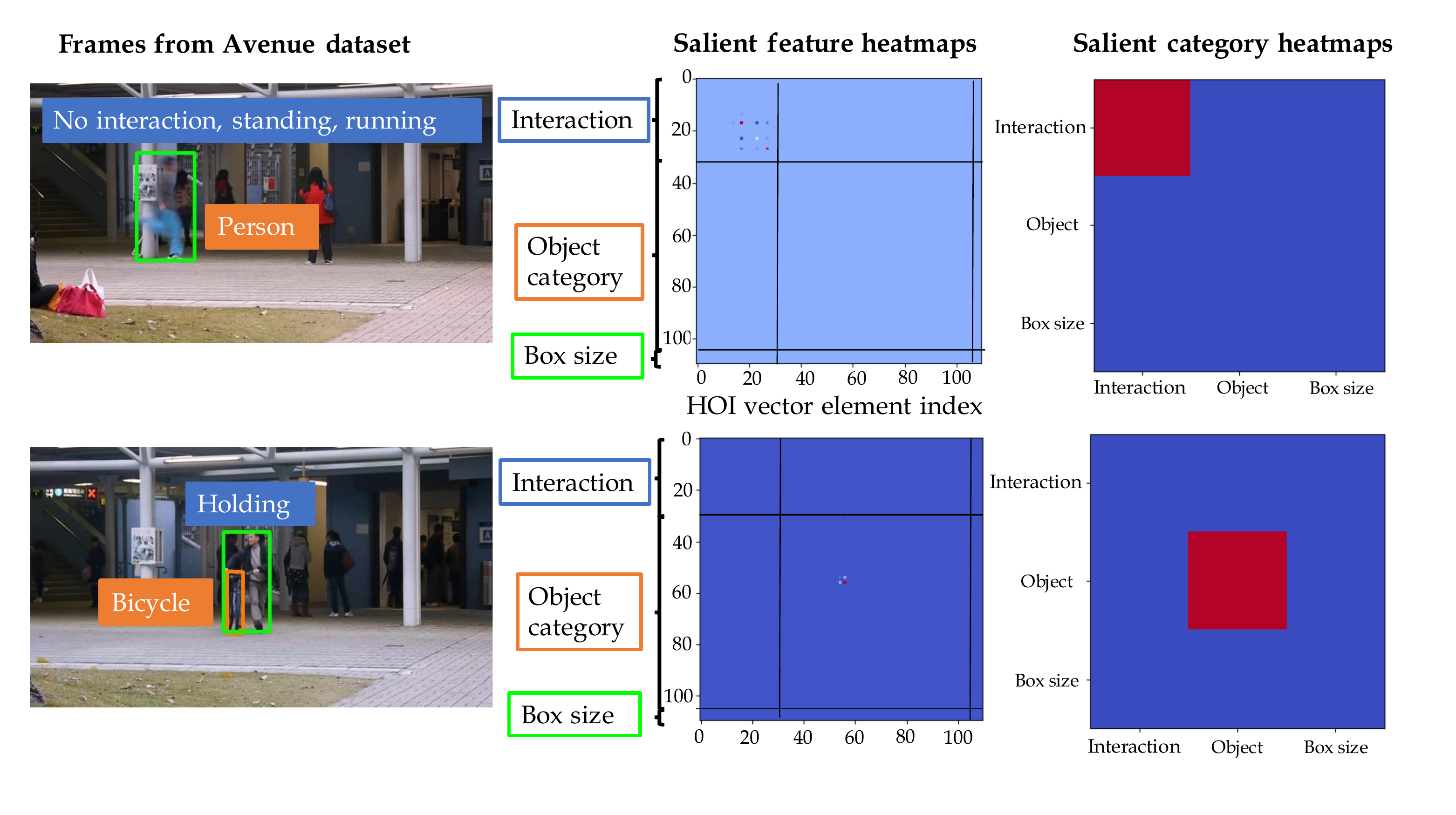}
\vspace{-10mm}
\caption{
\textbf{Heatmaps.} Examples of heatmaps for identification of salient features created with our explanation framework (middle column). First 29 elements of the vector correspond to the interaction class, the following 81 to the object category, the last 2 to the box size. Red indicates which pairs of elements contributed the most to the weighted distance from the normal event. We also summarise the results in a 3x3 heatmap (right column) show the salient category. Examples of cases when the interaction class was anomalous (top) and when the object was anomalous (bottom).} \label{explanation_hm}
\vspace{-5mm}
\end{figure*}

\paragraph{Saliency mapping.} Expressing the exponential component of equation~\ref{eq:deviation} using the original HOI vectors $\boldsymbol{v}$:
\begin{equation}
\begin{split}
& \boldsymbol{\Delta_{m}}^{T} \Sigma_{m}^{-1} \boldsymbol{\Delta_{m}}  = (\boldsymbol{x}-\boldsymbol{\mu}_{k})^{T} \Sigma_{k}^{-1} (\boldsymbol{x}-\boldsymbol{\mu_{k}}) \\ 
& = \sum_{i, j} (\boldsymbol{v}-\boldsymbol{y}_{k})_{i} (W \Sigma_{k}^{-1} W^{T})_{ij} (\boldsymbol{v}-\boldsymbol{y}_{k})_{j} 
\end{split}
\end{equation}
allows us to decompose the contribution of the summation (deviation from normal) across each element of the HOI vector, where $i$ and $j$ index the element. Each term in the summation consists of a difference between a pair of elements from the original HOI vector and normal event HOI vector $\boldsymbol{y} = W^{-1}\boldsymbol{\mu}$. These distances are weighted by $(W \Sigma_{k}^{-1} W^{T})_{ij}$ and visualised as a 2D heatmap. `Hot spots' being the feature with most contribution to the anomaly. The 2D location of the hot spots can explain which part of the HOI vector is responsible for the anomaly detection, e.g. if the hot spots lie in the upper left corner then the interaction class is responsible, yet if the hot spots spread along from the diagonal in the middle part of the vector, then the anomaly is due to an unexpected object , see Fig.\ref{explanation_hm} for examples. A 3x3 heatmap can also be formed which better visualises the contribution from object, interaction and bounding box size by taking the sum within each component, see Fig.\ref{explanation_hm}.

Quantitatively, we use the values on the diagonal of the heatmap as scores to indicate the classes responsible for the HOI being flagged as anomalous. To obtain per-frame explanation scores, we keep the maximum score of each class across all HOI vectors in a frame. The explanation score for location being anomalous is taken by summing the scores corresponding to the bounding box dimensions. The sum of scores across all classes is not normalised, hence allowing for multiple correct explanations (a frame can contain more than anomalous event). 

\section{X-MAN dataset \label{sec:dataset}}
We collect a new dataset for evaluating anomaly explanations: X-MAN (eXplanations of Multiple sources of ANomalies). The dataset supplements existing public anomaly detection datasets and provides labels describing the reason behind an anomalous event. Together, the existing public anomaly detection datasets and X-MAN, can be used as a benchmark for anomalous event localisation, i.e. detecting and categorising anomalies in video, therefore providing a measure of joint anomaly detection and explanation.

\subsection{Labelling anomaly explanations} 
\begin{figure}
\begin{centering}
\includegraphics[width=\linewidth]{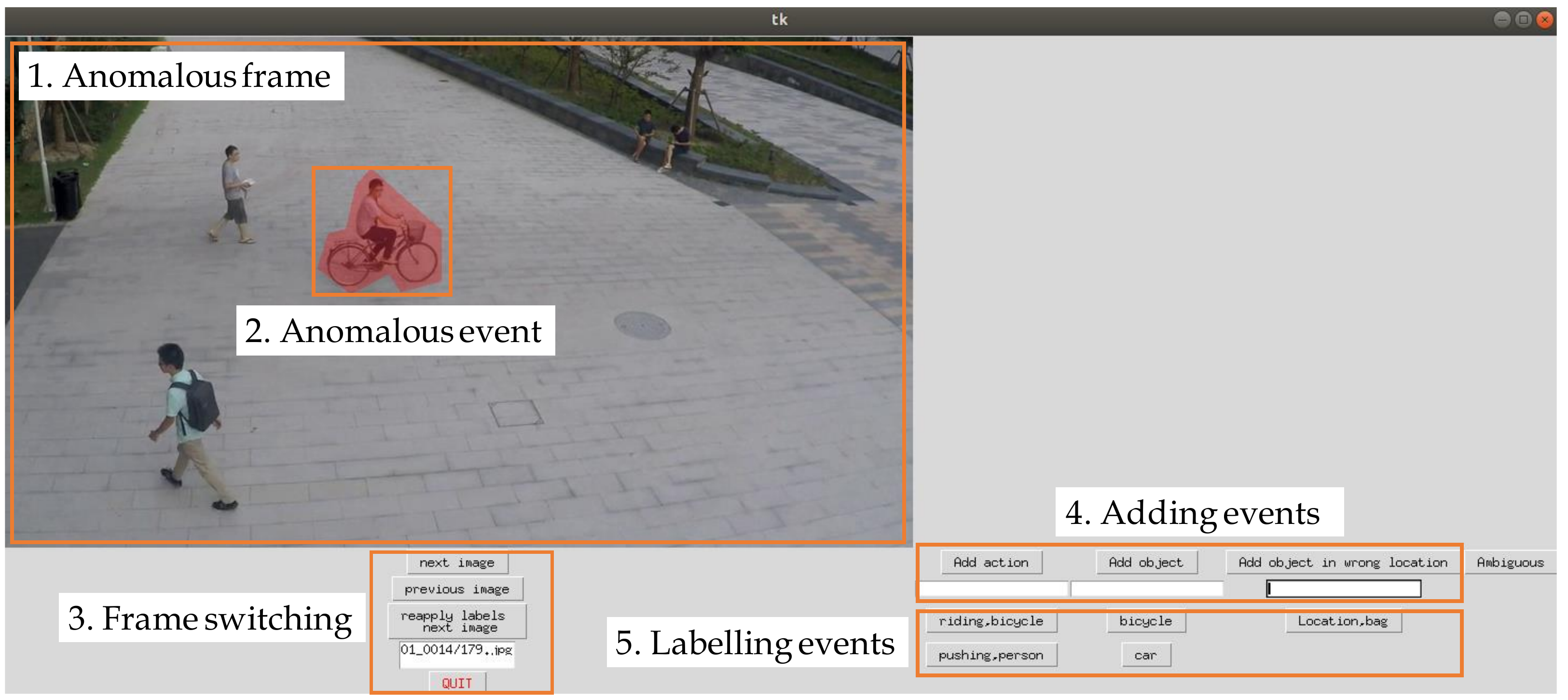}
\par \end{centering}
\caption{\textbf{Labelling tool.} GUI displays the anomalous frames in the video and provides tools to label the explanations. The GUI has 5 distinct areas: (1) highlights the abnormal region frame, (2) allows for switching frames and reapplying labels from the previous frame, (3) provides interface to add anomalous object and action categories which are show in (4), and finally (5) creates corresponding buttons that are used to label the frame with explanations.
} \label{fig:labelling_tool}
\end{figure}

The labelling task was conducted by human operators using a custom-built labelling tool (see Fig.~\ref{fig:labelling_tool}). The tool was used to label each anomalous frame in existing public datasets with explanations describing each of the anomalous events occurring in the frame on both a coarse level (\eg  action) and a fine-grained level (\eg running). Anomalous events are seen only in testing, hence this dataset can be used for evaluation only.

\paragraph{Coarse labels.} There are 3 possible classes of coarse labels: ``object", ``action" and ``location". They capture the high-level reason behind an event being categorised as anomalous. Only these 3 classes of coarse labels are considered, because they cover all anomalies in the public anomaly datasets and any obvious real-life examples (\eg guns or fighting).

\paragraph{Fine-grained labels.} Fine-grained labels explain the anomalous events in detail, supplementing the coarse labels in a hierarchical manner. In the case of a coarse ``object" label, the fine-grained label would specify which object made the frame anomalous, \eg bicycle. For the coarse ``action" label, fine-grained labels specify both the activity and the interaction object if one exists: examples include ``run", ``ride,motorcycle", ``fight/hit,person". In case of actions with more than one object involved, \eg person hitting another person with a bag, the interaction is decomposed into more than one interaction, each involving one object, \eg ``hit,person", ``hit,bag (with)". Finally, for the coarse "location" label, fine-grained labels specify what object is in an unusual location, \eg a human walking in a prohibited space or an unattended bag on the ground. 

Fine-grained object and action labels use categories defined in MS COCO~\cite{lin2014microsoft} and AVA Kinetics~\cite{li2020avakinetics}, respectively.

Both of these datasets capture a wide range of classes. However, it is expected that some anomaly categories will not be included, because anomalies are associated with rare events, hence they are unlikely to all be included in standard datasets. 

\subsection{Dataset statistics and metrics}

\paragraph{Statistics.} Our newly created dataset consists of 22,722 manually labelled frames in  ShanghaiTech (17,362), Avenue (3,712) and UCSD Ped2 (1,648). Each frame contains between 1 and 5 explanation labels, each label being a different reason why the frame is anomalous (many frames contain multiple anomalous events, \eg one person running and one riding a bike). In total, there are 40,618 labels across all frames. The majority of anomalies (22,640) are due to actions, followed by anomalous objects (14,828). The remaining anomalies are due to an anomalous location. 

\begin{figure}
\begin{centering}
\includegraphics[width = \linewidth]{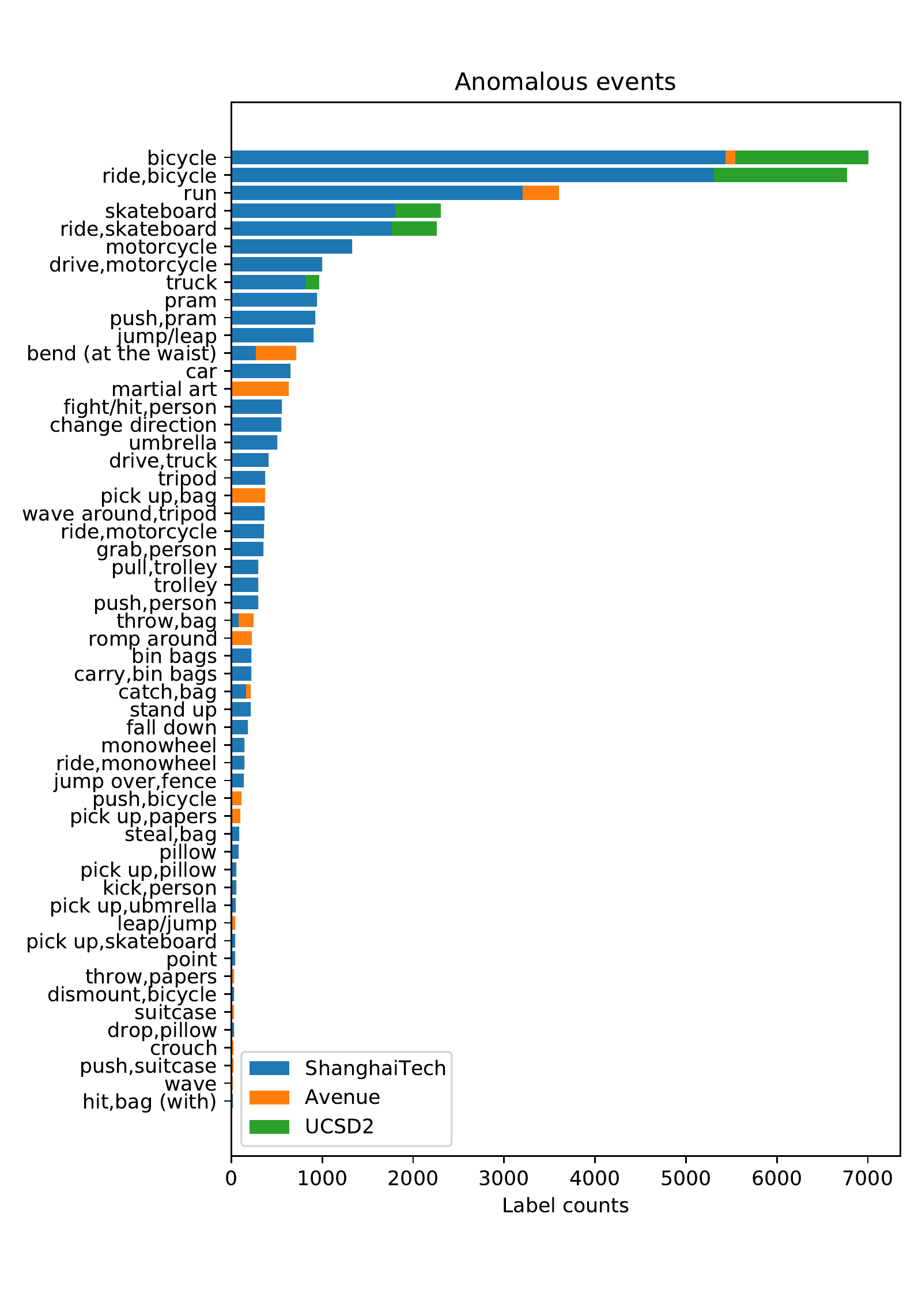}
\par\end{centering}
\caption{Distribution of anomalous actions across 22,000 labelled frames.\label{fig:action_statistics}}
\end{figure}

There are 42 anomalous actions (Fig.~\ref{fig:action_statistics}) 
and 13 anomalous objects. 
While the majority of actions can be found in AVA Kinetics, training videos with many action-object combinations in Fig.~\ref{fig:action_statistics} would not be found 
(because they are anomalous), and hence rare, making the X-MAN dataset challenging and complementary to existing ones.

\paragraph{Evaluation metric.} The task of explaining anomalies is similar to that of action recognition: in each anomalous frame the system has to recognise the anomalous event, there are multiple possible events in each frame and the recognition has to be accurate temporally. Hence, similar to action recognition, we propose to use mean Average Precision (mAP) as the evaluation metric for the anomaly explanation task. The mean is taken across the different explanation classes in order to weight rare explanation classes equally to the common ones.

\section{Experiments}
We evaluate our approach for both detecting anomalies and explaining them. For anomaly explanation we utilise our new dataset (described above), for anomaly detection we compare to state of the art on existing public datasets described below:

\subsection{Anomaly detection datasets and evaluation metric}

We evaluate our method on UCSD Ped2~\cite{mahadevan_2010}, Avenue~\cite{lu_2013} and ShanghaiTech~\cite{liu_2018}. 

\paragraph{UCSD.} A standard benchmark for anomaly detection. The training data contains only normal events, while testing data contains some abnormal events. 19600 frames captured using two different cameras: UCSD Ped1 and UCSD Ped2 which contains 16 training and 12 testing videos. Normal events include pedestrians walking, while abnormal events include trucks, cyclists and skateboarders. Following~\cite{hinami_2017} we evaluate on Ped2 only as Ped1 is very low resolution.

\paragraph{Avenue.} This dataset contains contains 16 training and 21 testing videos. All captured from the same scene, a total of 30,652 (15,328 training, 15,324 testing) frames. This is a challenging dataset because it includes a variety of events such as ``running", ``throwing bag", ``pushing bike" and ``wrong direction". We train from the videos in Avenue that contain normal events. This dataset focuses on dynamic events e.g. walking in an uncommon area in the scene and regards abnormal static events as normal e.g. standing in the same uncommon area.

\paragraph{Avenue17.} Was originally proposed by~\cite{hinami_2017} to better evaluate methods which can also detect static abnormal events. It consists of a subset of 17 videos from Avenue which exclude abnormal static events.

\paragraph{ShanghaiTech.} The largest and the most complex anomaly detection dataset. Contains 274,515 normal training frames in 330 videos and 42,883 testing frames in 107 videos with 130 complex abnormal events, \eg fighting, jumping over a fence. Videos are from 13 scenes with varying lighting and camera angles.

\paragraph{Evaluation metric.} All test video frames from all datasets are marked as either containing or not containing an anomaly. Measuring the true and false positive rates against this ground truth, we use the standard metric of evaluating abnormal event detection: the area under the ROC curve (AUC).

\begin{table*}[t!]
\begin{center}
\begin{tabular}{|l|l|l|l|l|}
\hline
Method & UCSD Ped2 & Avenue & Avenue17 & ShanghaiTech \\
\hline\hline
Kim et al. \cite{kim_2009} & 59.0 & - & - &  \\
Mahadevan et al. \cite{mahadevan_2010} &  82.9 & - & - & \\
Lu et al. \cite{lu_2013} & - & 80.6 & 80.3 &   \\
Hasan et al. \cite{hasan_2016} & 90.0 & 70.2 & 76.9 & 60.9 \\
Luo et al. \cite{luo_2017} & 92.2 & 81.7 & - & 68.0 \\ 
Liu et al. \cite{liu_2018} & 95.4 & 85.1 & - & 72.8\\
Park et al. \cite{park_2020} & 97.0 & 88.5 & - & 70.5 \\ 
\hline\hline
Hinami et al. \cite{hinami_2017}& 90.8 & - & 89.2 & - \\
\hline
Ours & 84.4 & 75.3 & 81.6 & 70.4  \\
\hline
\end{tabular}
\end{center}
\caption{Abnormal event detection accuracy in AUC (\%).\label{tab:results}}
\end{table*}

\begin{figure*}[t]
\includegraphics[width=\textwidth]{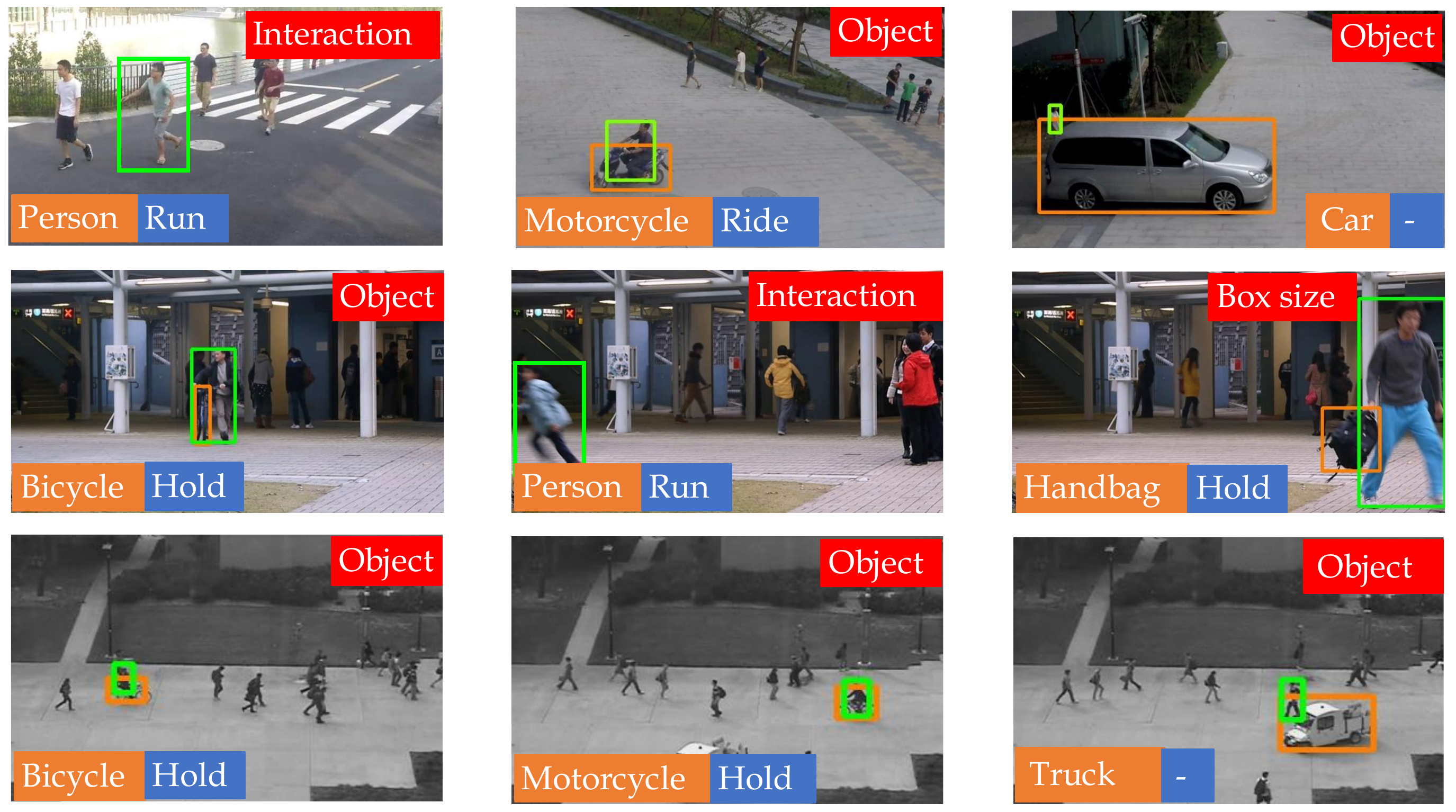}
\caption{\textbf{Correct detections.} Example anomalies detected in ShanghaiTech (top row), Avenue (middle row) and UCSD Ped2 (bottom row). Boxed areas show a human (green) object (orange) pair identified as anomalous. Red box in the top right corner indicates the salient feature causing the event to be detected as anomalous.} \label{all_tp}
\end{figure*}

\subsection{Implementation details}
We use an implementation of the object detector and DRG network from the authors of~\cite{gao_2020_drg}, and the provided pre-trained models from COCO~\cite{lin_2014} and V-COCO~\cite{gupta2015visual} datasets. The networks are trained to detect 29 interactions and 81 different objects including humans, human-human interactions are allowed. Each HOI vector is therefore $29+81+2=112$ (the bounding box width and height of the human detection are included in the vector, because they carry information about proximity of the person to the camera, hence implicitly providing information about location). We apply PCA dimensionality reduction. The output size varies between 19 and 23, depending on the dataset, and is chosen to capture over 99 \% of variance. 
The number of mixture components $M$ for GMM fitting is chosen based on the Bayesian information criterion~\cite{schwarz_1978} (BIC) and estimating the `elbow' of the BIC \textit{vs} $M$ curve. This varies between 2 and 7 depending upon the dataset.

\subsection{Results}

\paragraph{Quantitative anomaly detection state-of-the-art (SOTA) comparison.} Table~\ref{tab:results} summarises the AUC on all datasets. Our method performs competitively against all other methods. 
We believe our scores are slightly lower than the SOTA because it uses simpler clustering methods. However, it is the only method providing full explanations of the anomaly detection decisions. In a practical setting this is advantageous, allowing human operators of monitor systems to decide on an appropriate responses. Alternatively, it might allow for grading of the alert level raised by the anomaly, i.e. anomaly due to a gun is more alarming than an anomaly due to a person jumping.

\begin{figure*}[t!]
\includegraphics[width=\textwidth]{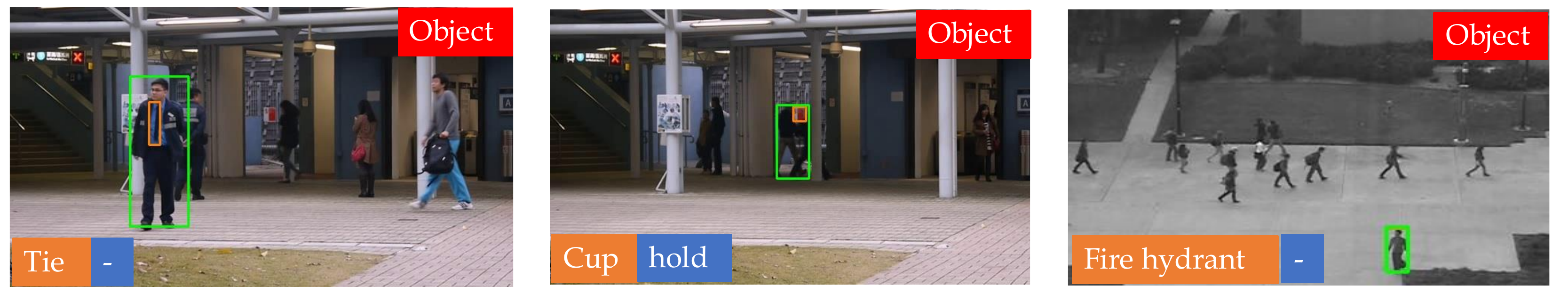}
\caption{\textbf{False positives.} Examples in Avenue and UCSD Ped2. Boxed areas show a human (green) object (orange) pair identified as anomalous. The left image is a false positive due to the failure of the anomaly detection system, while the other two are caused by a failure of the detector. Note how our explainability method helps identify the reason behind the models response i.e. incorrect object detection.} \label{fp}
\end{figure*}

\paragraph{Qualitative anomaly detection and explanation analysis.} Fig.\ref{all_tp}. shows examples of correctly detected anomalies from our method all 3 datasets. The interpreted anomaly is shown as an object and interaction pair (in orange and blue respectively). Using our saliency mapping method, we show in the red box the high level feature causing the event to be detected as anomalous. Examples of detected anomalies due to unexpected objects, interactions and box sizes are demonstrated. In Fig.\ref{fp}. examples of false positive failures of our anomaly detector are shown. Note how using high level features and the saliency mapping we can explain these failure cases quite easily i.e. incorrect object detection. Please also see our \textbf{supplementary video} showing the X-MAN dataset and anomaly detections and explanations from our system.

\paragraph{Quantitative anomaly explanation analysis.} Tab.~\ref{tab:explanation_results} shows the mAP achieved by our system on the anomaly explanation task. We evaluate our method against a baseline, where we remove the interaction part of our HOI vector, keeping only the object classes and bounding box size. The dataset contains classes that follow mostly the COCO and AVA datasets, but HOI methods, including DRG (which we use here), are trained on the V-COCO dataset, hence there is a mismatch of class labels. To mitigate this effect, we evaluate our method on the subset of classes in the X-MAN dataset that overlap with the classes in V-COCO and COCO datasets. In Tab.~\ref{tab:explanation_results} it can be seen that across all 3 datasets using interactions improves the explanation performance of our system, confirming our intuition that interactions are a crucial source of anomalies. 

\begin{table}
\begin{center}
\begin{tabular}{|l|l|l|l|l|l}
\hline
Method & UCSD & Avenue & Shanghai \\
 & Ped2 &  & Tech \\
\hline\hline
w/o interactions & 
21.2 & 
15.4 & 7.58 \\
\hline
with interactions & 
\textbf{42.1} & 
\textbf{17.2} & \textbf{11.7}\\
\hline
\end{tabular}
\end{center}
\caption{Abnormal event explanation mean Average Precision (mAP)
evaluated on the subset of X-MAN dataset, excluding interaction classes outside of V-COCO.  \label{tab:explanation_results}}
\vspace{-5mm}
\end{table}

\section{Conclusion}
This paper addresses the problem of explainable anomaly detection. For this task we release a large new dataset for evaluation and propose an interpretable probabilistic anomaly detection framework which can reason about high level video content. Our model automatically generates three levels of explanation for it's response: explaining the anomalous event, describing the closest normal event and highlighting the salient features. The qualitative analysis shows the potential of our explainability module and how it is able to indicate which of the interpretable features led to an anomaly. We also believe this is the first method for detecting anomalies using object interactions. Our model performs competitively against the state-of-the-art on 3 standard datasets. Quantitative evaluations on our newly gathered dataset show that using object interactions improves the anomaly explanations. Future work will focus on improving the interaction-object classifier on combinations not seen during the training phase, we believe this would help generalise the method further and increase performance of the anomaly detector.

{\small
\bibliographystyle{ieee_fullname}
\bibliography{references}
}

\end{document}